\def\year{2021}\relax
\documentclass[letterpaper]{article} 
\usepackage{aaai21}  
\usepackage{times}  
\usepackage{helvet} 
\usepackage{courier}  
\usepackage[hyphens]{url}  
\usepackage{graphicx} 
\usepackage{natbib}  
\usepackage{caption} 
\usepackage{amsmath}
\usepackage{amssymb}
\usepackage{mathtools}
\usepackage{amsthm}
\usepackage{enumitem}
\usepackage{subfig}
\usepackage{algorithm}
\usepackage[noend]{algpseudocode}
\usepackage{scrextend}

\newcommand{\spub}{s_{\mbox{\tiny pub}}}

\newcommand{\Spub}{\mathcal{S}_{\mbox{\tiny pub}}}
\newcommand{\Ipub}{I_{\mbox{\tiny pub}}}
\newcommand{\Opub}{O_{\mbox{\tiny pub}}}

\theoremstyle{definition}

\newcommand\blfootnote[1]{%
  \begingroup
  \renewcommand\thefootnote{}\footnote{#1}%
  \addtocounter{footnote}{-1}%
  \endgroup
}

\title{Solving Common-Payoff Games with Approximate Policy Iteration}
\author{
    Samuel Sokota,\textsuperscript{\rm 1}$^{\ast}$
    Edward Lockhart, \textsuperscript{\rm 2}$^{\ast}$\\
    Finbarr Timbers,\textsuperscript{\rm 2}
    Elnaz Davoodi,\textsuperscript{\rm 2}
    Ryan D'Orazio,\textsuperscript{\rm 3}\\
    Neil Burch,\textsuperscript{\rm 2}
    Martin Schmid,\textsuperscript{\rm 2}
    Michael Bowling,\textsuperscript{\rm 1, 2}
    Marc Lanctot \textsuperscript{\rm 2}
    \\
}
\affiliations{

    \textsuperscript{\rm 1}Universtiy of Alberta\\
    \textsuperscript{\rm 2}DeepMind\\
    \textsuperscript{\rm 3}Mila, Universit\'e de Montr\'eal\\
    sokota@ualberta.ca, 
    locked@google.com,\\ 
    finbarrtimbers@google.com,
    elnazd@google.com,
    ryan.dorazio@mila.quebec,\\
    burchn@google.com,
    mschmid@google.com,
    bowlingm@google.com,
    lanctot@google.com

}

\begin{document}

\maketitle

\begin{abstract}  
For artificially intelligent learning systems to have widespread applicability in real-world settings, it is important that they be able to operate decentrally.
Unfortunately, decentralized control is difficult---computing even an epsilon-optimal joint policy is a NEXP complete problem.
Nevertheless, a recently rediscovered insight---that a team of agents can coordinate via common knowledge---has given rise to algorithms capable of finding optimal joint policies in small common-payoff games.
The Bayesian action decoder (BAD) leverages this insight and deep reinforcement learning to scale to games as large as two-player Hanabi.
However, the approximations it uses to do so prevent it from discovering optimal joint policies even in games small enough to brute force optimal solutions.
This work proposes CAPI, a novel algorithm which, like BAD, combines common knowledge with deep reinforcement learning.
However, unlike BAD, CAPI prioritizes the propensity to discover optimal joint policies over scalability.
While this choice precludes CAPI from scaling to games as large as Hanabi, empirical results demonstrate that, on the games to which CAPI does scale, it is capable of discovering optimal joint policies even when other modern multi-agent reinforcement learning algorithms are unable to do so.
\end{abstract}

\maketitle

\section{Introduction}

In reinforcement learning~\cite{rlbook}, an agent seeks to learn a policy that extracts a large return from an environment, making it an algorithmic formulation of choice for many control problems.
However, the standard reinforcement learning framework assumes that decision making is centralized.
In general, a control problem may require multiple decision makers to act in a setting that bars complete information sharing, often in practice because decision makers are spatially separated.
This problem is known as decentralized control and has been studied by a number of different communities using varying formalisms \cite{dec-control,dec-pomdp,markovgames}, which we collectively refer to as \textit{common-payoff games}~\cite{gametheory}.\blfootnote{$^{\ast}$These authors contributed equally to this work.}\footnote{\textit{Cooperative game} is sometimes used as an umbrella term for the same purpose. This work uses \textit{common-payoff game} to avoid conflation with cooperative game theory, in which \textit{cooperative game} carries a distinct meaning.}

Despite involving multiple agents, the decentralized control problem bears relatively little similarity to general game-theoretic settings \cite{generalsum}, in which agents possess adversarial incentives and there is not a generally agreed upon notion of optimality.
It is also distinct from other problem settings operating under common-payoff game formalisms, such as ad-hoc coordination \cite{adhoc}, in which some of the agents are externally specified, or emergent communication \cite{emergent-comm}, in which coordination must arise naturally (not from precoordinated learning procedures).
Instead, in the decentralized control problem, the entity seeking to maximize return specifies all agents, including possibly precoordinated learning procedures.

While the decentralized control problem bears resemblance to the classical reinforcement problem in that both involve maximizing expected return, decentralized control presents challenges that do not arise in classical reinforcement learning.
Most notably, dynamic programming is not directly applicable to agents in decentralized control problems because the value of an agent's information state depends on the policies of its teammates.
One way to circumvent this issue is by alternating maximization \cite{alternating-maximization}.
Each agent maximizes its policy in turn, holding the policies of its teammates fixed.
This procedure guarantees convergence to a Nash equilibrium, but can be arbitrarily far away from an optimal joint policy, as measured by expected return.
Independent reinforcement learning (IRL) \cite{irl}, a paradigm in which all agents concurrently execute reinforcement learning algorithms, is another approach to dealing with decentralized control problems that has been the subject of attention in the deep multi-agent reinforcement learning community.
However, the convergence guarantees of IRL are less well-understood than those of alternating maximization and those that do exist show convergence only to local optima \cite{pg-convergence,ql-convergence}.

Arguably the most exciting line of research over the last decade in the pursuit of optimal joint policies for common-payoff games stems from the seminal insight of \citeauthor{nayyar}~\cite{nayyar}.
In their work, they show that, by conditioning on common knowledge, a team of decentralized agents can effectively act as a single agent, allowing for the direct application dynamic programming and resolving the difficulty of joint exploration.
But conditioning on common knowledge is not a panacea.
Finding an optimal, or even $\epsilon$-optimal, joint policy for a Dec-POMDP \cite{dec-pomdp}, the standard formalism for common-payoff games, is a NEXP-complete problem \cite{complexity,epsilon-complexity}.
While \citeauthor{nayyar}'s insights lead to solution methods for toy games, they are not immediately applicable to larger games, as conditioning on common knowledge is an exponential reduction.
And while there has significant progress in scaling similar ideas in the Dec-POMDP community \cite{dibangoye,dibrl}, resulting algorithms have largely been restricted to games having hundreds of states and fewer than ten actions.

At the time of writing, the Bayesian action decoder (BAD)~\cite{bad} is the only attempt that has been made to scale \citeauthor{nayyar}'s common knowledge approach to very large settings.
BAD is a policy-based approach relying on deep learning and independence assumptions.
When combined with population based training, BAD achieves good performance in two-player Hanabi.
However, while BAD achieves its intent of scalability, the approximations it requires to do so are so compromising that it struggles to solve games small enough to brute force an optimal solution.

Ideally there would an exist a common-knowledge approach matching or exceeding BAD's performance on large games that was also capable of solving small and medium-sized games.
This work takes a first step toward this goal by introducing \underline{c}ooperative \underline{a}pproximate \underline{p}olicy \underline{i}teration (CAPI), a novel deep approximate policy iteration algorithm for common-payoff imperfect information games.
Like BAD, CAPI seeks to scale \citeauthor{nayyar}'s insights.
But unlike BAD, CAPI prioritizes recovering the optimal joint policy over scalability.
To demonstrate the efficacy of CAPI, we consider two common-payoff games from OpenSpiel.
Having as many as tens of thousands of states and as many as hundreds of actions, these games are two orders of magnitudes larger than the common-payoff games that are often considered in Dec-POMDP literature \cite{dibangoye}.
We show that CAPI is able to achieve strong performance, solving (discovering an optimal policy of) both games a majority of the time.

\section{Background}

\subsection{Public and Common Knowledge}
Common knowledge has long been a subject of investigation in philosophy \cite{plato,convention}, multi-agent systems \cite{ck-distributed}, and (epistemic) game theory~\cite{Aumann76,sep-epistemic-game,Perea12Epistemic}.
Let $K_1^G$ be the set of information known to all agents in group $G$. 
Let $K_{i+1}^G$ be the subset of $K_i^G$ that is known by all agents to be in $K_i^G$.
Then \[K_{\mbox{\tiny common}}^G \coloneqq \cap_{i=1}^{\infty} K_i^G\]
is \textbf{common knowledge} among $G$.
The significance of common knowledge is that the inclusion status of any proposition is known by every member of the group.
In general, the same cannot be said for the set $K_i^G$ for any $i \in \mathbb{N}$.
This distinction has important implications regarding the abilities of groups of agents to coordinate their actions.

Unfortunately, while common knowledge appears to be essential for coordination in many settings, the infinite regress that defines it can make it expensive to compute \cite{mackrl,fog,cr,ck-phil,ck-distributed}.
In imperfect information games, a recent effort to circumvent this issue focuses attention on public knowledge, a special subset of common knowledge that is easily computable \cite{fog}.
Specifically, \textbf{public knowledge} is the subset $K_{\mbox{\tiny public}}^G \subset K_{\mbox{\tiny common}}^G$ of common knowledge that is publicly announced as such.

\subsection{Factored Observation Stochastic Games}

This work adopts finite common-payoff factored-observation stochastic game (FOSG) formalism \cite{fog}.
Finite common-payoff FOSGs have sufficient expressive power to represent finite Dec-POMDPs, the standard formalism for common-payoff games, and handle public knowledge in a principled manner.
A common-payoff FOSG is a tuple $G= \langle \mathcal{N}, \mathcal{W}, w^0, \mathcal{A}, \mathcal{T}, \mathcal{R}, \mathcal{O} \rangle$ where
\begin{itemize}[leftmargin=*]
    \item $\mathcal{N} = \{1, \dots, N\}$  is the \textbf{player set}.
    \item $\mathcal{W}$ is the set of \textbf{world states} and $w^0$ is a designated initial state.
    \item $\mathcal{A}= \mathcal{A}_1 \times \cdots \times \mathcal{A}_N$ is the space of \textbf{joint actions}.
    \item $\mathcal{T}$ is the \textbf{transition function} mapping $\mathcal{W} \times \mathcal{A} \to \Delta(\mathcal{W})$.
    \item $\mathcal{R}{=}(\mathcal{R}_1, \dots, \mathcal{R}_N)$ and $\mathcal{R}_1{=}\cdots{=}\mathcal{R}_N \colon \mathcal{W} \times \mathcal{A} \to \mathbb{R}$ is the \textbf{reward function}.
    \item $\mathcal{O} = (\mathcal{O}_{\mbox{\tiny priv}(1)}, \dots, \mathcal{O}_{\mbox{\tiny priv}(N)}, \mathcal{O}_{\mbox{\tiny pub}})$ is the \textbf{observation function} where
    \begin{itemize}[leftmargin=*]
    \item $\mathcal{O}_{\mbox{\tiny priv}(i)} \colon \mathcal{W} \times \mathcal{A} \times \mathcal{W} \to \mathbb{O}_{\mbox{\tiny priv}(i)}$ specifies the \textbf{private observation} that player $i$ receives.
    \item $\mathcal{O}_{\mbox{\tiny pub}} \colon \mathcal{W} \times \mathcal{A} \times \mathcal{W} \to \mathbb{O}_{\mbox{\tiny pub}}$ specifies the \textbf{public observation} that all players receive.
    \item $O_i{=}\mathcal{O}_i(w, a, w'){=}(\mathcal{O}_{\mbox{\tiny priv}(i)}(w, a, w'), \mathcal{O}_{\mbox{\tiny pub}}(w, a, w'))$ is player $i$'s \textbf{observation}.
    \end{itemize}
\end{itemize}
Note that actions need not be observable in FOSGs.

There are also a number of important derived objects in FOSGs.
\begin{itemize}[leftmargin=*]
    \item A \textbf{history} is a finite sequence $h=(w^0, a^0, \dots, w^t)$. We write $g \sqsubseteq h$ when $g$ is a prefix of $h$.
    \item The \textbf{set of histories} is denoted by $\mathcal{H}$.
    \item The \textbf{information state} for player $i$ at $h = (w^0, a^0, \dots, w^t)$ is $s_i(h) \coloneqq (O_i^0, a_i^0, \dots, O_i^t)$.
    \item The \textbf{information state space} for player $i$ is \\ $\mathcal{S}_i \coloneqq \{s_i(h) \mid h \in \mathcal{H}\}$.
    \item The \textbf{legal actions} for player $i$ at $s_i$ is denoted $\mathcal{A}_i(s_i)$.
    \item A \textbf{joint policy} is a tuple $\pi = (\pi_1, \dots, \pi_N)$, where \textbf{policy} $\pi_i$ maps $\mathcal{S}_i \to \Delta(\mathcal{A}_i)$.
    \item The \textbf{public state} at $h$ is the sequence \\$\spub(h) \coloneqq \spub(s_i(h)) \coloneqq (O_{\mbox{\tiny pub}}^0, \dots, O_{\mbox{\tiny pub}}^t)$.
    \item The \textbf{public tree} $\Spub$ is the space of public states.
    \item The \textbf{public set} for $s \in \Spub$ is $\Ipub(s){\coloneqq}\{h \mid \spub(h) = s\}$.
    \item The \textbf{information state set} for player $i$ at $s \in \Spub$ is $\mathcal{S}_i(s) \coloneqq \{s_i \in \mathcal{S}_i \mid \spub(s_i) = s\}$.
    \item The \textbf{reach probability} of $h$ under $\pi$ is \\$P^{\pi}(h) = P_{\mathcal{T}}(h) \prod_{i \in \mathcal{N}} P_i^{\pi}(h)$ where
    \begin{itemize}[leftmargin=*]
        \item Chance's contribution is \\$P_{\mathcal{T}}(h) \coloneqq \prod_{h'aw \sqsubseteq h} \mathcal{T}(h', a, w)$.
        \item Player $i$'s contribution is \\$P^{\pi}_i(h) \coloneqq P^{\pi}_i(s_i(h)) \coloneqq \prod_{s_i' a \sqsubseteq s_i(h)} \pi_i(s_i', a)$.
    \end{itemize}
\end{itemize}

\subsection{Public Knowledge in Common-Payoff Games} \label{sec:nayyar}

\begin{figure}
    \centering
    \includegraphics[width=\linewidth]{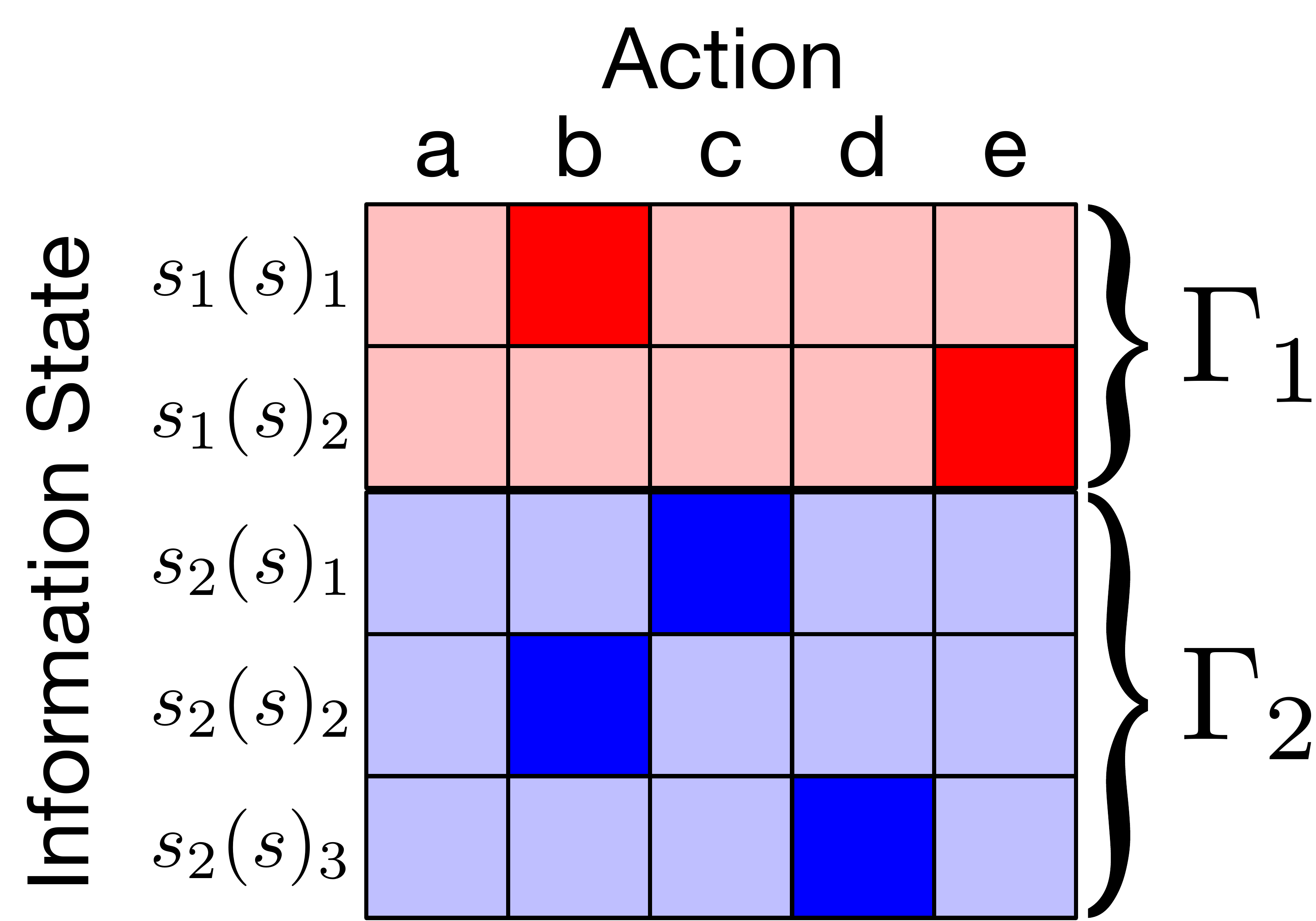}
    \caption{An example prescription vector. There are two players, with five actions each. Player one (red) has two possible information states while player two (blue) has three. The prescription vector decides each player's action as a function of its information state, as shown by the darkened squares.}
    \label{fig:prescription}
\end{figure}

\citeauthor{nayyar}~\cite{nayyar} were the first to formalize the general importance of public knowledge for coordinating teams of agents in common-payoff games.
They introduce \textit{the partial history sharing information structure}, a model for decentralized stochastic control resembling common-payoff FOSGs in that it explicitly acknowledges public observations.
\citeauthor{nayyar} show that this structure can be converted into a POMDP, which we refer to as the public POMDP.\footnote{In control literature, this is called the common information approach.}
Given a common-payoff FOSG $\langle \mathcal{N}, \mathcal{W}, w^0, \mathcal{A}, \mathcal{T}, \mathcal{R}, \mathcal{O} \rangle$, we can construct a \textbf{public POMDP} $\langle \tilde{\mathcal{W}}, \tilde{w}^0, \tilde{\mathcal{A}}, \tilde{\mathcal{T}}, \tilde{\mathcal{R}}, \tilde{\mathcal{O}} \rangle$ as follows.
\begin{itemize}[leftmargin=*]
    \item The world states of the public POMDP $\tilde{\mathcal{W}}$ are the histories $\mathcal{H}$ of the common-payoff FOSG.
    \item The initial world state of the public POMDP $\tilde{w}^0$ is the one tuple $(w^0)$.
    \item The actions of the public POMDP are called \textbf{prescription vectors}. A prescription vector is denoted by $\Gamma$ and has $N$ components.
    The $i$th component of a prescription vector $\Gamma_i$ is the \textbf{prescription} for player $i$. A prescription $\Gamma_i$ maps $s_i$ to an element of $\mathcal{A}_i(s_i)$ for each $s_i \in \mathcal{S}_i(\spub(h))$.
    In words, a prescription instructs a player in the common-payoff FOSG how to act as a function of its private information.
    An example is shown in Figure \ref{fig:prescription}.
    \item Given $\tilde{w} \equiv h$ and $\Gamma$, the transition distribution $\tilde{\mathcal{T}}(\tilde{w}, \Gamma)$ is induced by $\mathcal{T}(h, a)$, where \[a \equiv \Gamma(h) \coloneqq \left(\Gamma_1(s_1(h)), \dots, \Gamma_N(s_N(h))\right).\]
    \item Given $\tilde{w} \equiv h$ and $\tilde{w}'\equiv h'$, the reward $\tilde{\mathcal{R}}(\tilde{w}, \Gamma, \tilde{w}') \equiv \mathcal{R}_1(h, \Gamma(h), h') = \cdots = \mathcal{R}_N(h, \Gamma(h), h')$.
    \item Given $\tilde{w} \equiv h$ and $\tilde{w}' \equiv h'$, the observation $\tilde{\mathcal{O}}(\tilde{w}, \Gamma, \tilde{w}') \equiv \mathcal{O}_{\mbox{\tiny pub}}(h, \Gamma(h), h')$.
\end{itemize}

In summary, the public POMDP can be informally described as involving a \textit{coordinator} who only observes public observations and instructs the players how to act based on these observations.
One can imagine that this formulation could be executed at test time in a decentralized fashion by having each player carry around an identical copy of the central coordinator.
Since each player feeds its copy of the coordinator the same public observations, each copy of the coordinator produces the same prescription vector and it is as if a single coordinator were acting in the public POMDP.

\begin{figure}
    \centering
    \includegraphics[width=\linewidth]{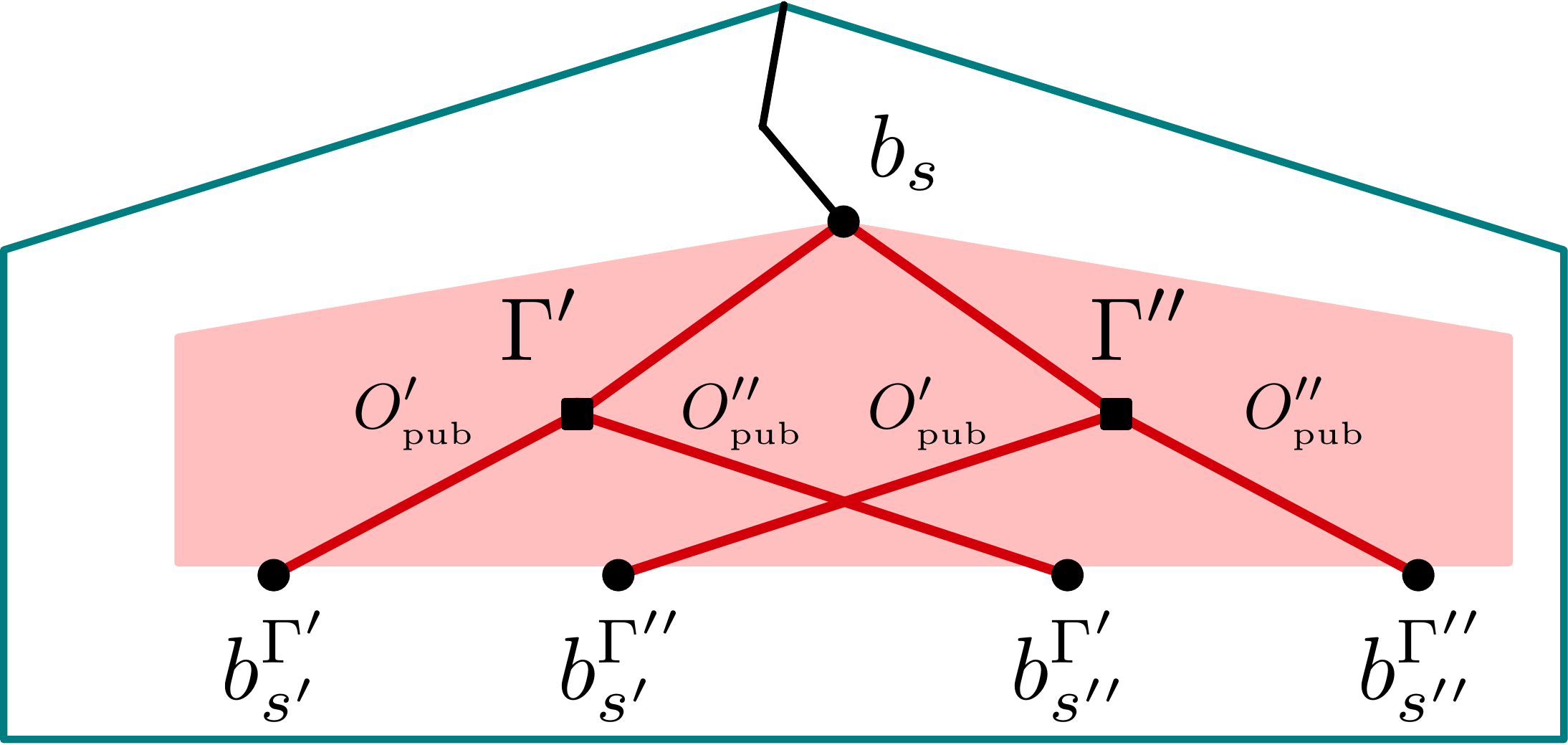}
    \caption{A visualization of a decision point in the PuB-MDP. The game begins in state $b_s$, a distribution over the public set $\Ipub(s)$. The coordinator is given a choice between two possible prescriptions $\Gamma'$ and $\Gamma''$. Both choices generate observations $O_{\mbox{\tiny pub}}'$ and $O_{\mbox{\tiny pub}}''$ with positive probabilities (inducing public states $s'$ and $s''$ respectively).
    Accordingly, there are four possible belief states for the next time step.}
    \label{fig:dlpsg}
\end{figure}

As with any with POMDP, the public POMDP can also be considered as a belief MDP, as is exampled in Figure \ref{fig:dlpsg}.
We follow the precedent set by \citeauthor{bad}, who refer to this perspective as the \textbf{public belief MDP} (PuB-MDP). 

The public POMDP and PuB-MDP have proven highly valuable and have been applied extensively in control literature \cite{app-ts,app-stateest,app-remoteest,app-meanfield,app-minimax,app-infhorizon,app-struct,app-suff,app-stoch,app-olig,phsis-mcp,reviewer-app,top-app} and, to a lesser extent, reinforcement learning \cite{bad,cirl,app-rl} literature.
Unfortunately, the public POMDP is so massive (there are roughly $|\mathcal{A}_i|^{N \cdot |\mathcal{S}_i|}$ actions at each decision point) that it is infeasible to apply POMDP solution methods \cite{recurrentdqn,mcp,despot,hsvi,joelle-pomdpsearch,Ross,perseus,Shani13POMDPs}, out-of-the-box, to common-payoff games of non-trivial size.

\section{The Tiny Hanabi Suite}

\begin{figure}[H]
    \centering
    \includegraphics[width=\linewidth]{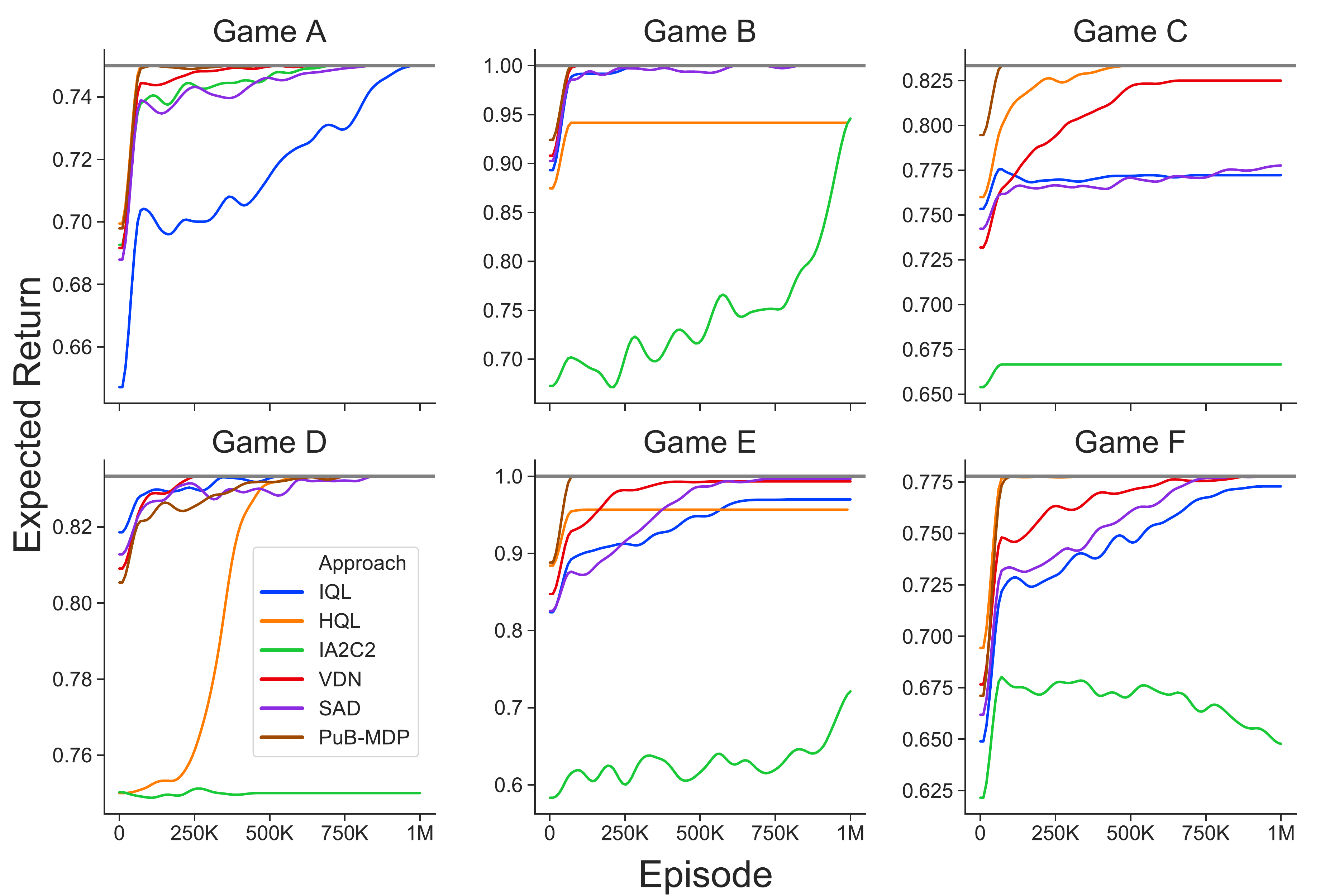}
    \caption{Performance comparison in the Tiny Hanabi Suite. Q-learning in the PuB-MDP consistently solves every game. In contrast, the algorithms operating within the independent reinforcement learning paradigm (IQL, HQL, IA2C2, VDN, SAD) are unable to do so.}
    \label{fig:th}
\end{figure}

To emphasize the contrast in guarantees between algorithms running in the PuB-MDP and independent learning algorithms, we display results from the Tiny Hanabi Suite---a collection of six very small Dec-POMDPs detailed in the appendix.
We compare independent Q-learning (IQL), hysteretic Q-learning (HQL) \cite{hql}, independent advantage actor centralized critic (IA2C2) \cite{maddpg,coma}, value decomposition networks (VDN) \cite{vdn}, simplified action decoding (SAD) \cite{sad}, and Q-learning in the PuB-MDP.
All algorithms were implemented tabularly and tuned across nine hyperparameter settings.
The results shown in Figure \ref{fig:th} are averages over 32 runs.
Code for the Tiny Hanabi Suite is available at \url{https://github.com/ssokota/tiny-hanabi}.

Despite tuning, reinforcement learning algorithms perform poorly.
HQL, VDN and SAD solve only four of the six games, IQL solves only three, and IA2C2 solves only one.
In contrast, Q-learning in the PuB-MDP solves all six games.
There results reaffirm that the differences in guarantees between algorithms that operate in the PuB-MDP and algorithms that operate within the independent reinforcement learning paradigm are not just theoretical---they also manifest in practice, even for modern multi-agent reinforcement learning algorithms.

\section{Cooperative Approximate Policy Iteration}

In this section, we introduce CAPI, a novel instance of approximate policy iteration operating within the PuB-MDP.
At a high level, at each decision point, CAPI generates a large number of prescription vectors using a policy and evaluates each of these prescription vectors according to its expected reward and the expected estimated value of the next belief state, selecting the most highly assessed prescription vector as its action.
We provide pseudocode for CAPI in Algorithm \ref{alg:cas}.
Each step is explained in greater detail below.
\begin{enumerate}[leftmargin=*]
\item At each decision point, CAPI takes a belief state $b$ and a number of rollouts $K$ as argument.
\item CAPI produces its policy $\pi(b)$ (over prescription vectors) as a function of the belief state. CAPI can either keep a separate tabular policy for each public state in the game or produce it by passing the belief state through a neural network.
\item CAPI acquires $K$ prescription vectors as a function of the policy $\pi(b)$.
Acquisition can be done either by sampling or taking the $K$-most-likely.
\item CAPI evaluates each of the $K$ prescription vectors. For each prescription vector $\Gamma^{(k)}$, this involves
\begin{enumerate}[leftmargin=*,nolistsep]
    \item Computing expected reward $r^{(k)}$ for $\Gamma^{(k)}$ given $b$.
    \item Computing the next belief state $b^{(k, \Opub)}$ for each $\Opub$.
    \item Estimating the value $v^{(k, \Opub)}$ of $b^{(k, \Opub)}$ using the value network.
    \item Computing the probability distribution $p^{(k)}$ over public observations given $b$ and $\Gamma^{(k)}$.
\end{enumerate}
The assessed value is the expected reward plus the expected estimated value of the next belief state
    \[q^{(k)} \gets r^{(k)} + \underset{\Opub \sim p^{(k)}}{\mathbb{E}} v^{(k, \Opub)}.\]
\item CAPI trains the policy to more closely resemble the most highly assessed prescription vector~$\Gamma^{(k^{\ast})}$ and the value network to more closely resemble the corresponding value~$q^{(k^{\ast})}$.
\item CAPI returns a random prescription vector among those it assessed if it is exploring. Otherwise it returns the most highly assessed prescription vector.
\end{enumerate}

\begin{algorithm}[H]
\caption{CAPI}
\label{alg:cas}
\begin{algorithmic}
\Procedure{act($b$, $K$)}{}
\State $[\Gamma^{(k)}] \gets \mbox{prescription\_vectors}(\pi(b), K)$
\State $[r^{(k)}] \gets \mbox{expected\_rewards}(b, [\Gamma^{(k)}])$
\State $[b^{(k, \Opub)}] \gets \mbox{next\_beliefs}(b, [\Gamma^{(k)}])$
\State $[v^{(k, \Opub)}] \gets \mbox{estimate\_values}([b^{(k, \Opub)}])$
\State $[p^{(k)}] \gets \mbox{pub\_observation\_probabilities}(b, [\Gamma^{(k)}])$
\State $[q^{(k)}] \gets [r^{(k)}]  + \mbox{pub\_expectation}([p^{(k)}], [v^{(k, \Opub)}])$
\State $k^{\ast} \gets \mbox{argmax}([q^{(k)}])$
\State $\mbox{add\_to\_buffer}(b, \Gamma^{(k^{\ast})}, q^{(k^{\ast})})$
\If {$\mbox{explore}$} 
\State \Return $\mbox{random}([\Gamma^{(k)}])$
\EndIf
\State \Return $\Gamma^{(k^{\ast})}$
\EndProcedure
\end{algorithmic}
\end{algorithm}

We run CAPI without sampling transitions, meaning that the episode is played out for every transition (i.e., every branch of the public tree) that occurs with positive probability, rather than sampling each transition.
After each episode, we train the value network and policy and wipe the buffer.

As of yet, we have left two important details unexplained.
First, how can CAPI pass a belief state over an exponential number of histories into a network?
To do so, CAPI adopts a trick from DeepStack \cite{DeepStack,valuefunctions}, and instead passes the public state $s$ and each player's contribution to the reach probability
\[\left[ P^{\pi}_i |_{\mathcal{S}_i(s)}(\cdot \mid s)\right]_{i \in \mathcal{N}}\]
as input, where $P^{\pi}_i$ denotes player $i$'s contribution to the reach probability and $f \mid_X$ denotes the function $f$ restricted to the domain $X$.
This information is a sufficient statistic for the belief state, but is more compact (albeit still exponential in the general case).
An explanation of sufficiency is offered in the appendix and can also be found in \citeauthor{valuefunctions}.

Second, how can CAPI maintain a policy over an exponentially large space?
To do so, CAPI adopts a trick from BAD \cite{bad} and uses a distribution over prescription vectors that factors across information states
\[P(\Gamma \mid b) = \prod_{i \in \mathcal{N}} \prod_{s_i \in \mathcal{S}_i(b)} \pi(\Gamma(s_i) \mid b).\]
This parameterization reduces the space required to store the distribution from $|\mathcal{A}_i|^{N \cdot |\mathcal{S}_i|}$ to $|\mathcal{A}_i| \cdot N \cdot |\mathcal{S}_i|$, making it explicitly manageable in the games we consider.
While this parameterization is constraining in that direct optimization by gradient ascent will only guarantee a local optima, search gives CAPI the opportunity to escape these local optima.
CAPI can optionally exploit this parameterization to add structured exploration by randomly setting rows of the policy to uniform random.

\section{Experiments}

\subsection{Problem Domains}

We consider two common-payoff games from OpenSpiel~\cite{openspiel} to demonstrate the efficacy of CAPI.
\subsubsection{Trade Comm}
The first is a communication game based on trading called Trade Comm.
The game proceeds as follows.
\begin{enumerate}[leftmargin=*]
    \item Each player is independently dealt one of \texttt{num\_items} with uniform chance.
    \item Player 1 makes one of \texttt{num\_utterances} utterances, which is observed by player 2.
    \item Player 2 makes one of \texttt{num\_utterances} utterances, which is observed by player 1.
    \item Both players privately request one of the $\texttt{num\_items} * \texttt{num\_items}$ possible trades.
\end{enumerate}
The trade is successful if and only if both player 1 asks to trade its item for player 2's item and player 2 asks to trade its item for player 1's item.
Both players receive a reward of one if the trade is successful and zero otherwise.
In our experiment, we use $\texttt{num\_items}=\texttt{num\_utterances}=12$.
This means that there is exactly enough bandwidth for the players to losslessly communicate their items and an optimal policy receives an expected reward of one.

This deceivingly easy-sounding game nicely illustrates the difficulty of common-payoff games.
It is too large to be tackled directly by using a public POMDP transformation and POMDP solution methods (the combination of which have been applied to games having fewer than 10 actions, whereas Trade Comm has 100s).
But simultaneously, as is shown in the appendix, independent deep reinforcement learning \cite{dqn,a3c} catastrophically fails to learn a good policy.

The code used to generate the results for CAPI is available at \url{https://github.com/ssokota/capi}.

\subsubsection{Abstracted Tiny Bridge}
For our second game, we consider Abstracted Tiny Bridge, which is a small common-payoff version of contract bridge retaining some interesting strategic elements.
In the game, each player is dealt one of 12 hands as a private observation.
The two players then bid to choose the contract.
The common payoff is determined by the chosen contract, the hand of the player who chose the contract, and the hand of player who did not chose the contract.
The challenge of the game is that the players must use their bids (actions) both to signal their hands and to select the contract, for which there are increasingly limited options as more bids are made.
The exact rules are detailed on OpenSpiel.

Despite its name, Abstracted Tiny Bridge is much larger than games traditionally considered in Dec-POMDP literature, having over 50,000 nonterminal Markov states.
For reference, Mars Rover \cite{mars-rover}, the largest game considered by many Dec-POMDP papers, has only 256.

\subsection{Baselines}

We again use tabular IQL, HQL, IA2C2, VDN, and SAD as baselines.
In our preliminary experiments the tabular implementations of these algorithms outperformed the respective deep variants.

\begin{figure}
    \centering
    \includegraphics[width=\linewidth]{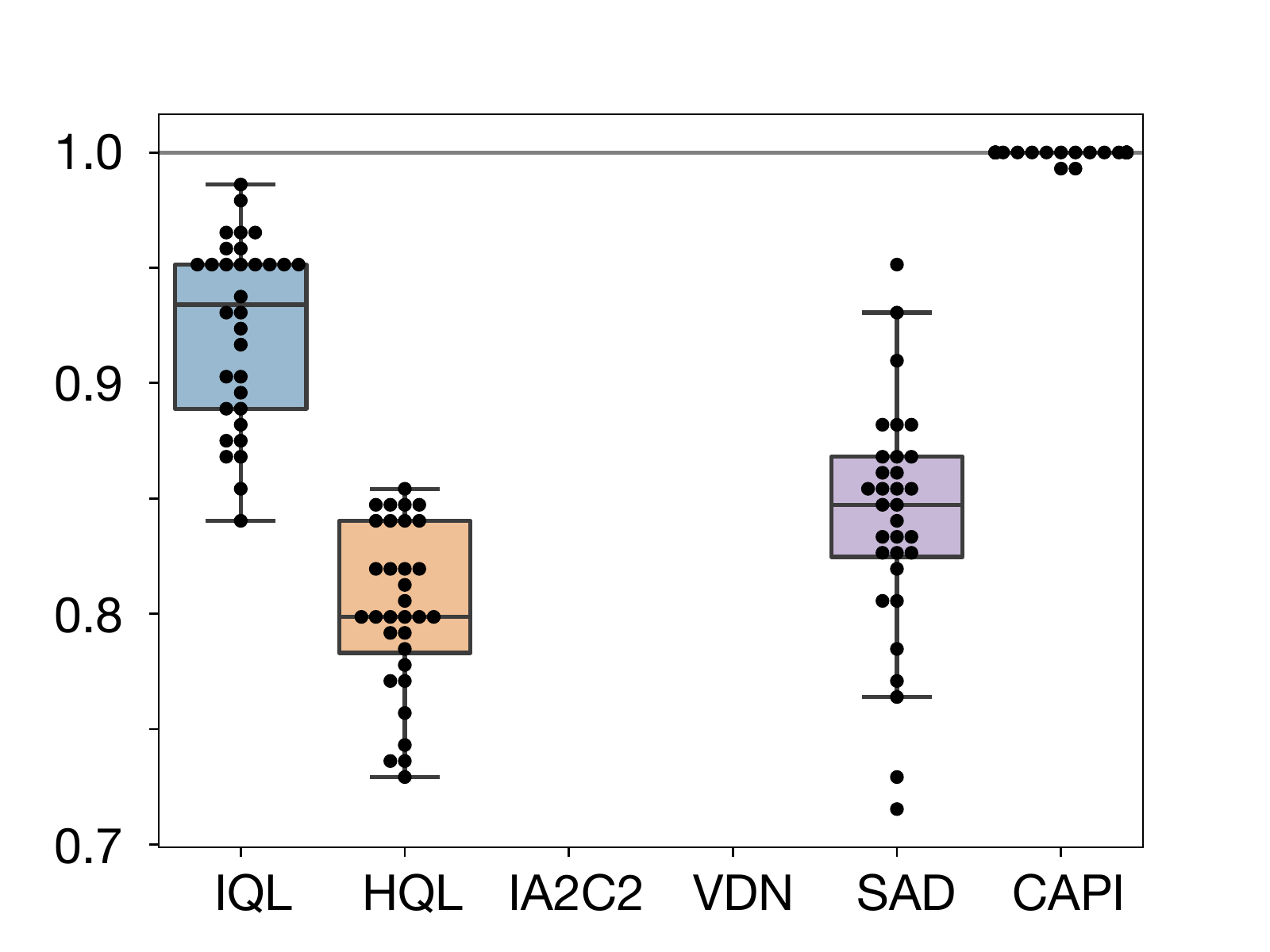}
    \caption{Performance comparison in Trade Comm. 
    CAPI consistently solves the game (30/32), whereas none of IQL, HQL, IA2C2, VDN, or SAD found an optimal policy on any of the 32 runs.
    Results for IA2C2 and VDN are below the bottom cutoff of the graph.
    Reported results are for best joint policy discovered.}
    \label{fig:tc}
\end{figure}

\subsection{Results}

Below, we describe the high level takeaways of our experiments.
Implementation details can be found in the appendix.

\subsubsection{Trade Comm}
In Figure \ref{fig:tc}, we show results for IQL, HQL, IA2C2, VDN, and SAD after 24 hours (as many as 100 million episodes) and results for CAPI after 2,000 episodes.
To give context to the scores on the graph, the optimal return is one (gray line) and the best joint policies that do not require coordination achieve an expected return of only 1/144.
IA2C2 and VDN (below bottom cutoff of graph) barely outperformed the best no-coordination joint policy.
The poor performance for IA2C2 may be attributable to the stochasticity of its policies, which are not well suited to Trade Comm's sparse rewards.
For VDN, the poor performance may be caused by its assumption that the joint value function decomposes additively, which provides a bad inductive bias for Trade Comm's non-monotonic value landscape.
In contrast to IA2C2 and VDN, HQL, IQL, SAD, and CAPI significantly outperform the best no-coordination joint policies.
Among the three, CAPI does the best, solving the game in 30 out of the 32 runs and nearly solving it on the remaining 2.
In contrast, none of IQL, HQL, and SAD solve Trade Comm on any of their respective 32 runs.

\subsubsection{Abtracted Tiny Bridge}
\begin{figure}
    \centering
    \includegraphics[ width=\linewidth]{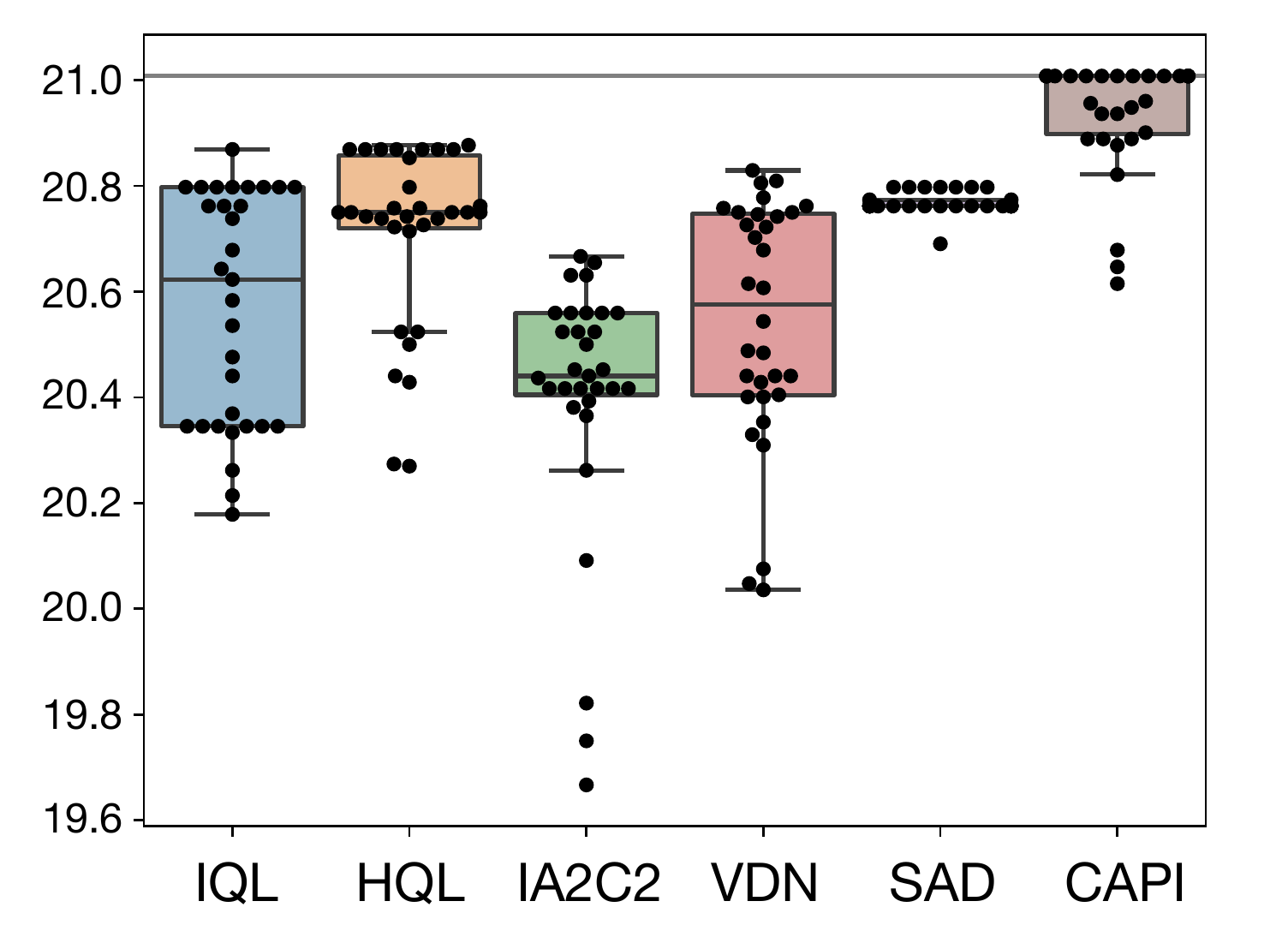}
    \caption{Performance comparison in Abstracted Tiny Bridge. CAPI solves the game on 18 of the 32 runs. None of IQL, HQL, IA2C2, VDN, or SAD solve the game on any of their respective runs. Reported results are for best joint policy discovered.}
    \label{fig:tiny-bridge-tabular}
\end{figure}

In Figure \ref{fig:tiny-bridge-tabular}, we show results for IQL, HQL, IA2C2, VDN, and SAD after 10 million episodes and results for CAPI after 100 thousand episodes.
To give context to the scores on the graph, the optimal return is the gray line and the best joint policy that we are aware of that does not require coordination achieves an expected return of 20.32.
IA2C2 arguably performed the worst, having both the lowest scoring minimum, median and maximum.
IQL and VDN performed comparably, with VDN performing slightly worse.
HQL and SAD arguably performed the best among the independent reinforcement learning algorithms, consistently discovering policies that outperform the others on average.
CAPI shows the strongest performance, solving the game on 18 of its 32 runs, contrasting the independent reinforcement learning algorithms, which do not solve the game on any of their runs and perform significantly worse on average.

\section{Related Work} \label{sec:relatedwork}

\subsection{From Dec-POMDP Literature}

Independently from \citeauthor{nayyar}, \citeauthor{dibangoye}~\cite{dibangoye} show that a common-payoff game can be converted into a non-observable MDP (NOMDP) \cite{oli-nomdp}, a special kind of POMDP in which the agent (in this case the coordinator) receives no observations.
\citeauthor{dibangoye} call the corresponding belief MDP the occupancy state MDP. 
\citeauthor{dibangoye} also introduce feature-based heuristic search value iteration (FB-HSVI), a novel planning algorithm with asymptotic optimality guarantees.
In a large-scale study, \citeauthor{dibangoye} show that combining the occupancy state, FB-HSVI, and equivalence relations over information states is able to solve games with hundreds of states and outperforms contemporaneously existing methods.

\citeauthor{GT} build on the occupancy state MDP approach, showing that it can be extended to infinite horizon problems when beliefs are bounded.
They also show how belief compression can be combined with point-based value iteration \cite{perseus} to achieve good empirical results.
There is also recent work that extends the occupancy state MDP approach to the reinforcement learning problem setting \cite{dibrl} and to special cases involving asymmetric observability \cite{new-dib}.

Compared to the games investigated in these works, the games investigated in this work are larger.
However, the scalability that is gained by leveraging deep learning and CAPI's policy parameterization precludes CAPI from having the same guarantees as these algorithms.

\subsection{From Cooperative MARL Literature}

Outside of this work, there has also been work by the multi-agent reinforcement learning community to apply the PuB-MDP at scale.
BAD \cite{bad} scales a policy-based approach to the PuB-MDP by using an independence assumption across private features to analytically approximate belief states and an independence assumption to parameterize the distribution over prescription vectors.
\citeauthor{bad} also introduce a belief update correction procedure resembling expectation propagation \cite{ep} and use population-based training to show how BAD can be scaled to two-player Hanabi.

Compared to BAD, CAPI is significantly less scalable---two-player Hanabi is much larger than the games considered in this work.
However, the cost of this scalability is that BAD cannot solve games even as small as game E of the Tiny Hanabi Suite.
In contrast, our results demonstrate that CAPI is capable of discovering optimal policies.

There has also been work applying  decision-time search at scale in common-payoff games.
SPARTA \cite{sparta} runs a one-ply search over the next action in the game at each public belief state, using an externally specified policy as a blueprint according to which the rest of the rollout is played.
\citeauthor{sparta} show that in the limit in the number of rollouts, SPARTA produces a policy no worse than its blueprint.
In practice, empirical results on Hanabi suggest that SPARTA's policy tends to significantly outperform its blueprint.

Similarly to SPARTA, CAPI performs decision-time planning in the PuB-MDP.
Unlike SPARTA, CAPI is a self-contained learning algorithm and does not require an externally specified blueprint.
It is also unlike SPARTA in that its design principles are geared toward solving games, whereas SPARTA is designed for scalability and has failure cases even in small games.

\subsection{From Adversarial Game Literature}

Public belief states have also played an important role in recent successes in adversarial games.
In poker, DeepStack \cite{DeepStack}, Libratus \cite{libratus}, and Pluribus \cite{pluribus} combine public belief states \cite{safenested,cr} with equilibrium computation methods and precomputed value functions \cite{valuefunctions}.
More recently, \citeauthor{rebel} proposed ReBeL \cite{rebel}, which is similar to CAPI in that both learn policies and value functions over public belief states from self-play and use decision-time planning.

\subsection{From Mixed-Motive Game Literature}

Recent work on the game Diplomacy is similar to CAPI in how it handles a combinatorial action space \cite{dm-diplomacy, fair-diplo}.
The sampled best response algorithm \citeauthor{dm-diplomacy} propose is especially similar in that it also performs a one-ply search using a variant of policy iteration with sampled actions.
The equilibrium search algorithm \citeauthor{fair-diplo} propose is also similar in that it performs a one-ply search over the $K$-most-likely actions.

\section{Conclusion and Future Work}

In this work, inspired by BAD \cite{bad} and the seminal work of \citeauthor{nayyar}, we propose CAPI, a novel approximate policy iteration algorithm for common-payoff imperfect information games.
In many ways, CAPI is similar to BAD, but differs notably in that it is more correct and less scalable.
Empirically, we demonstrate the efficacy of CAPI by showing that it is able to solve games on which modern multi-agent reinforcement learning algorithms struggle.

The direction for future work that we are most interested in is the construction of a single public knowledge approach inheriting CAPI's performance (or better) on small games and BAD's performance (or better) on large games. 
The existence of such an algorithm would be an exciting and unifying discovery.
It may require effective mechanisms for learning or approximating public belief states and for working with compressed or implicit representations of prescription vectors.
Both of these mechanisms would also be of significance for two-player zero-sum games, where the intractability of public belief states and prescription vectors is a limiting factor for applying decision-time planning approaches to larger problems.

\section{Acknowledgements}

\begin{itemize}
    \item We thank Thomas Anthony and Arthur Guez for providing feedback that substantially improved the quality of this work.
    \item The writing and content of this work are based on and overlap with \cite{thesis}.
\end{itemize}

\bibliography{capi}

\section{Compact Belief Factorization}
\label{app:belief-factorization}
In general, the size of the public set grows exponentially in time.\footnote{Though in practice, there are many cases in which the public set does not grow exponentially.
One example is in games in which all private observations occur at the beginning, such as Texas Hold'em or Stratego.
Another example is in games in which old private observations are revealed when new private observations are introduced, such as Hanabi.
A third is settings in which imperfect recall is sound, as is true in some graphical security games.}
Even assuming that the public set is small enough to keep a posterior over, this posterior may still be too unwieldy for practical use as input to a value or policy network.
By exploiting playerwise factorization, we can produce a sufficient statistic that, while still growing exponentially, is much smaller than the posterior over histories \cite{DeepStack,valuefunctions}.
Observe that the publicly conditioned probability of any history can be expanded as
\[P^{\pi}(h \mid s)
= \frac{P_{\mathcal{T}}(h) \prod_{i=1}^N P^{\pi}_i(s_i(h))}{\sum_{h' \in \Ipub(s)} P_{\mathcal{T}}(h') \prod_{i=1}^N P^{\pi}_i(s_i(h'))}.\]
It follows that the publicly conditioned posterior over histories can be losslessly reconstructed from the public state and each player's contribution function by a mapping of the form
\[\left[s, P^{\pi}_1 |_{\mathcal{S}_1(s)}(\cdot \mid s), \dots, P^{\pi}_N |_{\mathcal{S}_N(s)}(\cdot \mid s)\right] \longmapsto P^{\pi} |_{\Ipub(s)}(\cdot \mid s)\]
that is independent of the historical policy profile.
Therefore, the input of this mapping is a sufficient to estimate the public belief state value.\footnote{Note that since the actions chosen by the coordinator induce a deterministic policy profile, the restriction of $P^{\pi}_{i}(\cdot \mid s)$ to $\mathcal{S}_i(s)$ is a binary vector with ones at $s_i \in \mathcal{S}_i(s)$ consistent with $\pi_i$ and zeros elsewhere.}
While this statistic still grows exponentially in the length of the game, it is a much smaller dimensionality than the posterior over histories and it grows only linearly in the number players.

\section{Independent Deep Reinforcement Learning for Trade Comm}
\label{app:trade-comm-irl-fail}

To emphasize the difficulty of Trade Comm, we ran a hyperparameter sweep for independent DQN and A2C. 
The DQN agents were trained with all combinations of following parameters: learning rate $[2e-1, 1e-1, 1e-2, 1e-3, 1e-4]$, decaying exploration starting at $[1., 0.8, 0.5, 0.2]$, hidden layer size of \{'64', '128', '32, 32', '64, 64'\} and replay buffer capacity of $[50, 100, 1000, 10000]$. 
The A2C agents were trained with all combinations of following parameters: critic learning rate $[1e-2, 1e-3]$, actor learning rate $[1e-3, 1e-4, 1e-5]$, number of critic updates before every actor update $[4, 8, 16]$ and hidden-layer size of \{'32', '64', '128', '32, 32', '64, 64'\}.
Both the independent DQN agents and the independent A2C agents were trained for $5e{+}6$ steps.
We ran 3 runs at each parameter setting.
In total the parameter sweep took 45 hours and 14 minutes.

We trained both the independent DQN agents and the independent A2C agents to play a version of Trade Comm with 10 items and 10 utterances. 
The best DQN agents achieved an expected return of around 0.12. 
No A2C agents achieved an expected return of more than 0.01.

\section{Implementation Details}
\label{app:hyperparams}

\subsection{IQL, HQL, IA2C2, VDN, SAD}

We used a linearly decaying epsilon and a linearly decaying learning rate for both games.

\subsubsection{Trade Comm}
For Trade Comm, the rate of decay was set such that both the epsilon value and the learning rate value would reach zero after 100 million episodes.
To determine the initial values, we did a grid search over the values $[0.1, 0.2, 0.3, 0.4, 0.5]$ for the initial epsilon and $[0.1, 0.2, 0.3, 0.4, 0.5]$ for the initial learning rate.

\subsubsection{Abstracted Tiny Bridge}
For Abstracted Tiny Bridge, the rate of decay was set such that both the epsilon value and the learning rate value would reach zero after 10 million episodes.
We did a grid search over the values $[0.1, 0.2, 0.3]$ for the initial epsilon and $[0.05, 0.1, 0.2]$ for the initial learning rate.

\subsection{CAPI}

In both games we used 3 layer neural network with 256 units per hidden layer and the Adam optimizer for the value function.

\subsubsection{Trade Comm}
For Trade Comm we used a two-headed network to produce the value and policy, randomly sampled 10,000 prescription vectors from the policy, and used $\epsilon$-greedy exploration with $\epsilon =0.1$.
We used a learning rate of $1e-4$, a weight of 1/100 for the loss of the policy head and a weight of 1 for that of the value head.
We used mean squared error for the value head and cross entropy loss for the policy head.

\subsubsection{Abstracted Tiny Bridge}

For Abstracted Tiny Bridge we maintained a tabular policy for each public state, added structured exploration by overwriting a random row of the policy to be uniformly random during training, generated the 10,000 most likely prescription vectors from the policy, and played an exploratory prescription vector once an episode.

For the policy we used a learning rate of $1e{-}6$ and a floor of 0.01.
We explored once every episode.
We also added structured exploration by setting a row of the prescription vector to uniform random at each decision point.
We squashed the output of the value network to be between -1 and 1 by applying sigmoid to the output.
We used a mean squared error loss for the value network.

\section{The Tiny Hanabi Suite}
The Tiny Hanabi Suite is a collection of six toy common-payoff games created by Neil Burch and Nolan Bard.
Each game is structured in three steps.
\begin{enumerate}[leftmargin=*]
\item A dealer samples two cards from separate piles of  \texttt{num\_cards} cards with uniform probability and gives the first card to player one and the second player two.
\item Player one chooses one of \texttt{num\_actions} actions. Player two observes player one's action.
\item Player two chooses one of \texttt{num\_actions} actions.
\end{enumerate}
After player two's action, the game ends and the common payoff is determined as a function of the cards dealt and the actions taken.

Although the tiny Hanabi games are most naturally thought of as temporally-extended, the payoff functions can be succinctly described as payoff tables, as is shown in Figure \ref{fig:gameA}, Figure \ref{fig:gameB}, Figure \ref{fig:gameC}, Figure \ref{fig:gameD}, Figure \ref{fig:gameE}, and Figure \ref{fig:gameF}.
\begin{itemize}[leftmargin=*]
    \item The uppercase numerals in the leftmost column represent the card dealt to player one.
    \item The lowercase numerals in the uppermost column represent the card dealt to player two.
    \item The uppercase letters represent player one's action.
    \item The lowercase letters represent player two's action.
    \item The integer corresponding to a particular quadruplet is the payoff to that trajectory.
\end{itemize}

In games A, B, C, and D, \texttt{num\_cards}$=2$ and \texttt{num\_actions}$=2$.
In game E, \texttt{num\_cards}$=2$ and \texttt{num\_actions}$=3$.
In game F, \texttt{num\_cards}$=3$ and \texttt{num\_actions}$=2$.

\subsection{Game A}
\begin{figure}[H]
    \centering
    \begin{tabular}{|l|l|rr|rr|}
    \hline
    Card &  & i &  & ii &  \\
     & Action & a & b & a & b \\
    \hline
    I & A & 0 & 1 & 0 & 1 \\
     & B & 0 & 0 & 3 & 2 \\
    \hline
    II & A & 3 & 3 & 2 & 0 \\
     & B & 3 & 2 & 3 & 3 \\
     \hline
    \end{tabular}
    \caption{Game A of the Tiny Hanabi Suite.}
    \label{fig:gameA}
\end{figure}

\subsection{Game B}
\begin{figure}[H]
    \centering
    \begin{tabular}{|l|l|rr|rr|}
    \hline
    Card &  & i &  & ii &  \\
     & Action & a & b & a & b \\
    \hline
    I & A & 1 & 0 & 0 & 1 \\
     & B & 1 & 0 & 0 & 1 \\
    \hline
    II & A & 0 & 1 & 1 & 0 \\
     & B & 0 & 0 & 1 & 0 \\
     \hline
    \end{tabular}
    \caption{Game B of the Tiny Hanabi Suite.}
    \label{fig:gameB}
\end{figure}

\subsection{Game C}
\begin{figure}[H]
\centering
\begin{tabular}{|l|l|rr|rr|}
\hline
Card &  & i &  & ii &  \\
 & Action & a & b & a & b \\
\hline
I & A & 3 & 0 & 2 & 0 \\
 & B & 0 & 3 & 3 & 3 \\
\hline
II & A & 2 & 2 & 0 & 1 \\
 & B & 3 & 0 & 0 & 2 \\
 \hline
\end{tabular}
\caption{Game C of the Tiny Hanabi Suite.}
\label{fig:gameC}
\end{figure}

\subsection{Game D}
\begin{figure}[H]
\centering
\begin{tabular}{|l|l|rr|rr|}
\hline
Card &  & i &  & ii &  \\
 & Action & a & b & a & b \\
\hline
I & A & 3 & 0 & 3 & 0 \\
 & B & 1 & 3 & 3 & 0 \\
\hline
II & A & 3 & 2 & 0 & 1 \\
 & B & 0 & 2 & 0 & 0 \\
 \hline
\end{tabular}
\caption{Game D of the Tiny Hanabi Suite.}
\label{fig:gameD}
\end{figure}

\subsection{Game E}
\begin{figure}[H]
\centering
\begin{tabular}{|l|l|rrr|rrr|}
\hline
Card &  & i &  &  & ii &  &  \\
 & Action & a & b & c & a & b & c \\
\hline
I & A & 10 & 0 & 0 & 0 & 0 & 10 \\
 & B & 4 & 8 & 4 & 4 & 8 & 4 \\
 & C & 10 & 0 & 0 & 0 & 0 & 10 \\
\hline
II & A & 0 & 0 & 10 & 10 & 0 & 0 \\
 & B & 4 & 8 & 4 & 4 & 8 & 4 \\
 & C & 0 & 0 & 0 & 10 & 0 & 0 \\
 \hline
\end{tabular}
\caption{Game E of the Tiny Hanabi Suite.}
\label{fig:gameE}
\end{figure}

\subsection{Game F}
\begin{figure}[H]
\centering
\begin{tabular}{|l|l|rr|rr|rr|}
\hline
Card &  & i &  & ii &  & iii &  \\
 & Action & a & b & a & b & a & b \\
\hline
I & A & 0 & 3 & 0 & 0 & 3 & 1 \\
 & B & 3 & 2 & 0 & 1 & 2 & 1 \\
\hline
II & A & 0 & 2 & 1 & 2 & 0 & 1 \\
 & B & 0 & 1 & 1 & 2 & 0 & 3 \\
\hline
III & A & 1 & 3 & 0 & 3 & 3 & 1 \\
 & B & 1 & 2 & 2 & 2 & 3 & 0 \\
 \hline
\end{tabular}
\caption{Game F of the Tiny Hanabi Suite.}
\label{fig:gameF}
\end{figure}

\end{document}